\documentclass{llncs}
\usepackage[OT1]{fontenc}
\usepackage{makeidx}  
\usepackage{times}  
\usepackage{helvet}  
\usepackage{courier}  
\usepackage{url}  
\usepackage{graphicx}  
\usepackage{latexsym}
\usepackage{comment}
\usepackage{tabu}  
\usepackage{color}
\usepackage{multirow}
\usepackage{amsmath} 
\DeclareMathOperator*{\argmax}{arg\,max}
\usepackage{amsfonts} 
\usepackage{subfigure}
\urldef{\mails}\path|{xwshi, hhy63, zsyprich, pjian, tangyk}@bit.edu.cn|
\makeatletter
\newcommand{\thickhline}{%
    \noalign {\ifnum 0=`}\fi \hrule height 1pt
    \futurelet \reserved@a \@xhline
}
\makeatother
\begin{document}
\frontmatter          
\pagestyle{headings}  
\addtocmark{Hamiltonian Mechanics} 
%
%

\title{Tag Recommendation by Word-Level Tag Sequence Modeling}
\author{Xuewen Shi\inst{1,2} \and
Heyan Huang\inst{1,2} \and
Shuyang Zhao\inst{1} \and
Ping Jian\inst{1,2}\thanks{Corresponding author} \and
Yi-Kun Tang\inst{1,2}}
\authorrunning{X. Shi et al.}
\institute{
School of Computer Science and Technology, Beijing Institute of Technology, \\
Beijing 100081, P.R. China \and
Beijing Engineering Research Center of High Volume Language Information Processing and Cloud Computing Applications
\mails\\}
\maketitle
\begin{abstract}
Recently, neural network based tag recommendation has gained much attention.
Most of these methods regard tag recommendation as multi-label classification.
They treat different tags as individual categories with fixed labels,
without considering the rich relations among tags.
Moreover, using a fixed number of categories cannot tackle dynamic, changing tags with ongoing topics.
In this paper, we transform tag recommendation into a word-based text generation problem and introduce a sequence-to-sequence model.
For efficiently modeling the semantic dependencies among tags in tag sequence and the strong sequential relations among the tag-words, we propose an essential sequence-to-sequence model, named LSTM-Attention.
The model inherits the advantages of recurrent network based encoder for sequential modeling and attention based decoder for learning relations globally.
In addition, as a text generation method, the proposed model is able to generate unseen tags, which is more applicable and flexible to real scenarios.
Extensive experimental results on two datasets, \textit{i.e.}, \textit{Zhihu} and \textit{Weibo}, clearly illustrate the proposed model significantly outperforms other state-of-the-art text classification based methods and
well demonstrate its advantage of handling unseen tags.
\end{abstract}
\section{Introduction}
\label{sec:intro}
In recent years, online Q\&A community and social network platform have become important modes for information transfer, such as Zhihu and Sina Weibo.
These corpora contain a form of metadata tags marking its keywords or topics.
These tags are useful in many real-world applications, \textit{e.g.,} information retrieval \cite{Efron2010hashtag,saha2013discriminative,bansal2015towards},
sentiment analysis \cite{buddhitha2015topic,kalamatianos2015sentiment},
hot issues discovery \cite{wang2016momentum,kim2018twitter},
public opinion analyses \cite{ma2016public,meduru2017opinion}, to name a few.
Therefore, automatic tag recommendation has gained a lot of research interests recently.
\begin{figure}[t]
\centering
\includegraphics[width=0.6\textwidth]{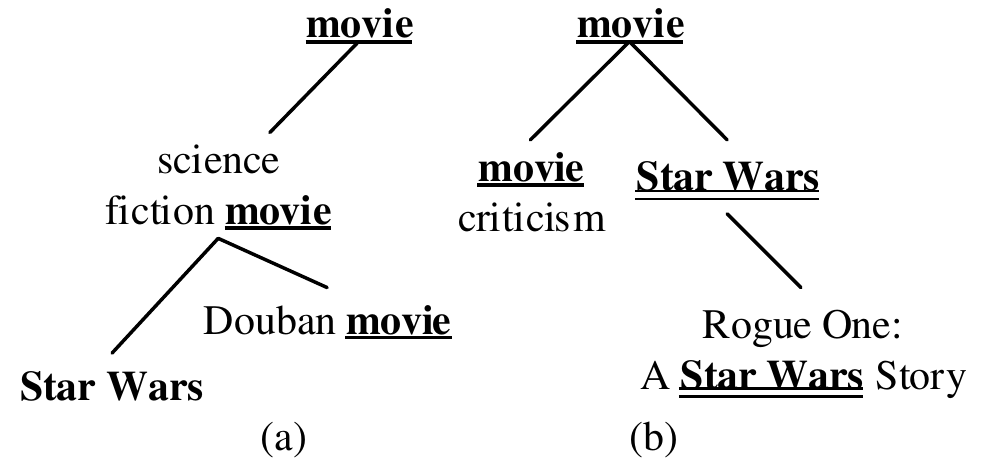}
\caption{
\textbf{Illustration of the internal relations between tags.}
These graphs show the semantic hierarchy between tags (from \textit{Zhihu} questions).
The \underline{underlined words} 
are the word-level common parts (called cues in this paper) shared by different tags.
}
\label{fig:introduction1}
\end{figure}

Most previous neural network based approaches regard tag recommendation as multi-label classification. They typically treat different tags as separate categories and learn the classifier with fixed labels, while very few consider the rich relations between tags. However, in practical applications, new labels arise with the emergence of new topics. Thus using a fixed number of labels cannot meet the actual needs.
Additionally, the labels tagged on a specific text are often related to each other.
This association includes the word-level similarity and the hierarchical semantic relations. As shown in Fig.~\ref{fig:introduction1}, related tags often share common words, \textit{e.g.,} the word ``movie'' is shared by ``movie'', ``science fiction movie'', and ``Douban movie'' tags. It can also be observed that there exist semantic relations between those tags in a certain text.
In the tag hierarchy in Fig. \ref{fig:introduction1}, ``movie'' and ``science fiction movie'' tags have a parent-child relationship. Obviously, previous approaches which learn different tags as separate categories cannot leverage such rich relations.
\begin{figure*}[t]
\centering
\includegraphics[width=0.65\textwidth]{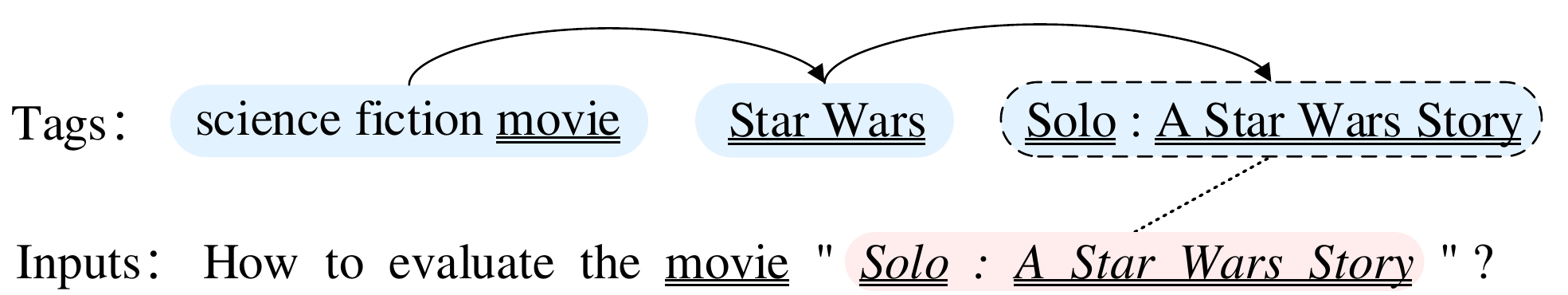}
\caption{
\textbf{Illustration of our motivation.} We use word-level cues to guide the model generating unseen tags.
The 
{\underline{underlined words}} in the ``Inputs'' are word-level cues.
The labels surrounded by dashed lines are unseen tags.
The directed edge refers to the ``belonging'' relation between two tags. See \S~\ref{sec:intro} for more detailed discussion.
}
\label{fig:introduction}
\end{figure*}

With above novel insights, we propose here a new word-based tag recommendation model, \textit{i.e.}, modeling this task as a word-based text generation problem. In many cases, a tag also shares common units with its relevant text (see the underlined words in Fig. \ref{fig:introduction}). We call the common parts as cues, for emphasizing their values of predicting accurate tags.
For classification based tag recommendation methods, the only cue that can be used is the full tag name. However, for our word-based model, the cues can be any single word that makes up the tags.
Table~\ref{tab:wordtextconnections} presents the ratio of the cues appearing in the full tags and the words of tags; showing that the word-level cues are more flexible and widespread. Additionally, leveraging those word-level cues allows our model to handle unseen tags.
Taking Fig.~\ref{fig:introduction} as an example, ``Solo: A Star Wars Story'' is a new tag which is unseen during the training step,
but its components ``Solo'' and ``A Star Wars Story'' which are word-level cues in the source input, have already been included in the datasets.
Thus it is feasible to predict the unseen tags by using those cues.

\begin{table}[ht]
  \centering
  \setlength\tabcolsep{10pt}
  \begin{tabular}{c||cc}
  \hline\thickhline
  Element &Zhihu &Weibo \\
  \hline
  \hline
  full tags &~~30.0\%~~ &~~6.1\%~~\\
  word of tags&~~38.2\%~~ &~~24.1\%~~\\
  \hline
  \end{tabular}
  \caption{\textbf{Comparisons of the coverage ratio between tag-level and word-level elements in inputs for different datasets}, where the word-level cues are more widespread.}
  \label{tab:wordtextconnections}
\end{table}

Our approach is achieved by a sequence-to-sequence model for fully exploring the internal relations between tags. Basically, it's built upon an encoder-decoder architecture,
where the encoder maps the input features into a sequence of continuous source representations,
and the decoder, which takes all previous predicted tags and the source representation sequence as the conditions simultaneously, generates the predicted tags word-by-word. There exist semantic dependencies among the tags in tags sequence, such as the tags ``movie'', ``science fiction movie'', and ``Douban movie'' in Fig.~\ref{fig:introduction1}.
On the other hand, there are strong sequential relations within tags, such as the words ``science'', ``fiction'', ``movie'' in the tag ``science fiction movie''.
For efficiently capturing above diverse relations in tag generation, we propose an essential LSTM-Attention model,
where the LSTM encoder is used for modeling the sequential relations within inputs,
while the attention model based decoder is designed for learning the semantic relations from a global view. The attention model is achieved by Transformer~\cite{vaswani2017attention}, which is further extended with a \textit{local positional encoding} strategy,
to disengage tags explicitly and couple the words from a same tag tightly.
In order to better utilize the semantic relations within tag sequences,
we propose two ordering strategies
to reorganize tag sequences
and quantitatively assess their effects,
offering a deeper insight into the suggested word-level tag generation method.


The proposed tag recommendation model has several interesting characteristics and advantages:
\begin{itemize}
\item It treats tags as word sequences and generates them without the limit of a fixed tag list. It learns the latent rules of grouping words into tags and thus is able to produce unseen tags. By the use of richer word-level cues, it gains improved generability and applicability.
\item An essential LSTM-Attention model is designed for modeling the semantic relations among tags globally and learning the sequential dependencies among the words from the same tags.
\item A local positional encoding strategy is integrated into the Transformer based decoder to address the complex relations in tag sequences.
\item To better leverage the semantics and regularize the generation order of tags, two ordering strategies of tag sequence are explored by considering the top-down and bottom-up relations between tags.
\end{itemize}

We evaluate our model on two tasks, \textit{i.e.}, open-data tagging in \textit{Zhihu} dataset and the multi-label classification in \textit{Zhihu} and \textit{Weibo} datasets.
The extensive experiments clearly demonstrate that our model achieves better or competitive performances compared with state-of-the-art methods, and show significant advantage on predicting unseen tags.
\section{Related Work}
\label{sec:related}

There have been proposed several neural network based tag recommendation approaches, most of which ~\cite{gong2013deep}
cast this problem as multi-label classification.
For example, Gong \textit{et al.}~\cite{gong2013deep} investigate ranking-based loss functions for CNN and proposed weighted approximated-ranking loss for multi-label annotation.
Huang \textit{et al.}~\cite{huang2016hashtag} use end-to-end memory networks to incorporate the histories of users with attention mechanism to select more appropriate histories.
Rawat \textit{et al.}~\cite{rawat2016contagnet} proposed a context-aware model based on CNN and Context Network.
Gong \textit{et al.}~\cite{gong2016hashtag} presented an attention based CNN method for tag recommendation.
Li \textit{et al.}~\cite{li2016tweet} use sentence vectors to train an LSTM model to classify tags.
Topical Attention-Based LSTM \cite{li2016hashtag} combines topic model with LSTM through attention mechanism.
Zhang \textit{et al.}~\cite{zhang2017hashtag} proposed a co-attention network incorporating textual and visual information to recommend hashtags for multimodal tweets.
Huang \textit{et al.}~\cite{huang2017multi} proposed a Bayesian model by exploiting local positive and negative pairwise label correlations.
Yeh \textit{et al.}~\cite{yeh2017learning} proposed a deep network based model to learn a feature-aware latent subspace for label embedding.
Zhu \textit{et al.}~\cite{zhu2018multi} proposed a new multi-label approach GLOCAL exploiting global and local label correlations simultaneous.

All of these works show promising results and well demonstrate the benefit in exploiting the deep learning architectures in this problem.
However, they seldom explore the rich semantic relations between the words that make up the labels and largely do not address the issue of generating new unseen tags.
In comparison, we build our tag recommendation approach upon a sequence-to-sequence word-level text generation model, which is able to efficiently explore the inherent semantics and rich cues in the problem. Additionally, through representing tags as word sequence, the proposed model is able to handle unseen tags.

\section{Method}

\subsection{Word-level Tag Prediction}
\label{sec:nmt}
We re-formalize tag recommendation as word-level text generation. 
Formally,
given a sequence of input source features $\mathbf{X}=\{\mathbf{x}_1,..,\mathbf{x}_N \}$,
our method seeks to find a tag word sequence $\mathbf{y}=\{y_1,..,y_T\}$ that maximizes the conditional probability of $\mathbf{y}$ given $\mathbf{X}$, \textit{i.e.},
$\argmax_{\mathbf{y}}{p(\mathbf{y}|\mathbf{X})}$.

We adopt an encoder-decoder~\cite{cho2014learning} architecture to tackle this sequence-to-sequence problem.
The model mainly consists of two components:~\romannumeral1)~an encoder which maps a source feature sequence $\mathbf{X}$ into
a sequence of continuous representations $\mathbf{Z}=\{\mathbf{z}_1,...,\mathbf{z}_N\}$;
and~\romannumeral2)~a fully attention based decoder that generates output sequence $\mathbf{y}$ one word at a time taking the input representations $\mathbf{Z}$ and the previous outputs as the conditions.
The decoder defines a probability over the output sequence $\mathbf{y}$ by decomposing the joint probability into the ordered conditionals:
\begin{equation}
p(\mathbf{y}|\mathbf{X})=\prod\nolimits_{t=1}^{T}{p(y_t|y_1,..,y_{t-1},\mathbf{Z})}.
\label{eq:jointrobability}
\end{equation}
Each conditional probability at decoder time step $t$ is modeled as:
\begin{equation}
p(y_t|y_1,..,y_{t-1},\mathbf{Z}) = \mathcal{G}(y_1,...,y_{t-1},\mathbf{Z}),
\end{equation}
where $\mathcal{G}(\cdot)$ is a neural network based implementation of the decoder.

\subsection{Model Architecture}
For modeling above tag sequence generation, we propose a novel sequence-to-sequence model, named LSTM-Attention.
Different from natural sentences, tag sequences involve rich semantic relations between tags and sequential dependencies between tag words.
But there is no obvious sequential dependency between tags (as discussed in \S~\ref{sec:intro}).
Therefore, in our encoder-decoder model, an LSTM encoder is designed to capture the sequential dependencies between tag-words via its recurrent nature,
and an attention model (Transformer \cite{vaswani2017attention}) based decoder is applied for learning semantic relations globally (instead of sequentially).
The LSTM encoder and the attention model are integrated as a powerful tag generation model for effectively mining the rich word-level cues, producing reasonable tags and handling unseen tags.

\noindent\textbf{LSTM-based Encoder.}
The LSTM based encoder is composed of a stack of two bi-directional LSTM layers.
Each bi-directional layer consists of forward and backward LSTMs.
For each layer $k$~$\in$~$\{1,2\}$, the forward LSTM $\overrightarrow{\mathcal{R}}^{k}$ reads the output from the prior LSTM in forward manner while the backward LSTM $\overleftarrow{\mathcal{R}}^{k}$ leverages the inputs in reverse order.
The the outputs from the forward and backward LSTMs are concatenated and fed into the next layer as input.

Formally, at step $n$, we calculate the output $\mathbf{z}^{k}_n$ of $k$-th encoder layer as follows:
\begin{equation}
\overrightarrow{\mathbf{z}}^k_n,\overrightarrow{\mathbf{c}}^k_n=\overrightarrow{\mathcal{R}}^{k}({\mathbf{i}^k_n},\overrightarrow{\mathbf{z}}^k_{n-\!1},\!\overrightarrow{\mathbf{c}}^k_{n-\!1}),
\label{eq:lstm1}
\end{equation}
\begin{equation}
\overleftarrow{\mathbf{z}}^k_n,\overleftarrow{\mathbf{c}}^k_n=\overleftarrow{\mathcal{R}}^{k}({\mathbf{i}^k_n},\overleftarrow{\mathbf{z}}^k_{n+\!1},\!\overleftarrow{\mathbf{c}}^k_{n+\!1}),
\label{eq:lstm2}
\end{equation}
\begin{equation}
\mathbf{i}^{k+1}_n=\mathbf{z}^{k}_n=
\begin{bmatrix}
\overrightarrow{\mathbf{z}}^k_n\\
\overleftarrow{\mathbf{z}}^k_n
\end{bmatrix},
\label{eq:lstm3}
\end{equation}
where $\mathbf{i}_n^k$ indicates the input to $\mathcal{L}^k$ at time step $n$. For the first LSTM layer, we set $\mathbf{i}_n^1 = \mathbf{x}_n$.
$\mathbf{z}^k_t$ and $\mathbf{c}^k_n$ are the hidden states and memory states, respectively. In Eq.~\ref{eq:lstm3}, the output $\mathbf{z}^{k}_n$ of $k$-th LSTM layer is used as the input $\mathbf{i}^{k+1}_n$ of the next layer.
Given the input feature sequence $\mathbf{X}=\{\mathbf{x}_1,..,\mathbf{x}_N \}$, the final output $\mathbf{Z}=\{\mathbf{z}^2_1,...,\mathbf{z}^2_N\}$ of the last LSTM encoder will be fed into an attention based decoder for tag sequence generation.

\noindent\textbf{Attention-based Decoder.}
The Transformer based decoder $\mathcal{G}(\cdot)$ is composed of four identical layers.
Each of the layers is composed of three stacked sub-layers,
including a multi-head self-attention, a multi-head attention over the output of the encoder stack, and a point-wise, fully connected feed-forward network.
Each sub-layer is equipped with residual connections, and followed by layer normalization.

An attention function can be described as mapping a query and a set of key-value pairs to an output.
The output is computed as a weighted sum of the values,
where the weight assigned to each value is computed by a function of the query with the corresponding key.
For multiple queries, the queries, keys and values are packed into matrices $\mathbf{Q}$, $\mathbf{K}$ and $\mathbf{V}$ respectively.
Vaswani \textit{et al.}~\cite{vaswani2017attention} adopted the ``Scaled Dot-Product Attention'' into Transformer.
It consists of queries and keys of the same dimension $d_K$. 
Then it computes the dot products of the query with all keys, divides each by $\sqrt{d_{k}}$, and applies a $softmax$ function to obtain the weights on the
values.
For multiple queries, the attention outputs can be calculated as:
\begin{equation}
\mathcal{A}(\mathbf{Q},\mathbf{K},\!\mathbf{V})\!=\!softmax(\frac{\mathbf{Q}\mathbf{K}^\top}{\sqrt{d_{K}}})\mathbf{V}.
\end{equation}

In a self-attention layer $\mathcal{A}(\cdot)$, all of the $\mathbf{Q}$, $\mathbf{V}$ and $\mathbf{K}$ come from the same place: the output of the previous layer in the
decoder.
In "decoder attention over the output of the encoder" layers, the $\mathbf{Q}$ comes from the previous decoder layer,
and $\mathbf{K}$ and $\mathbf{V}$ are from the output of the encoder stack $\mathbf{Z}$.
For more details about the multi-head self-attention layer and position-wise feed-forward networks, we refer the reader to~\cite{vaswani2017attention}.

\noindent\textbf{Local Positional Encoding.}
The transformer introduces positional encoding~\cite{vaswani2017attention} to make use of the order of the sequence.
The calculation functions for each dimension are as follows:
\begin{equation}
\begin{aligned}
\mathcal{L}(p,2c)&=sin(p/10000^{2c/d_{M}}), \\
\mathcal{L}(p,2c+1)&=cos(p/10000^{2c+1/d_{M}}),
\end{aligned}
\label{eq:positional_encoding}
\end{equation}
where $p$ is the position and $c$ is the dimension index. The model feature size $d_{M}$ is set to $512$.

For tag sequence, as we discussed before, there exist strong sequential relations among tag-words and rich semantic (instead of sequential dependencies) among tags.
Thus the above purely sequential location encoding is limited, as it fails to reflect the independence of tags within tag sequence.
To remedy this, we propose here a local positional encoding strategy.
The difference with Transformer's positional encoding is that the $p$ in Eq.~\ref{eq:positional_encoding} is the relative position of a word in each tag instead of the position in the whole sequence.
For example, `movie$|$science fiction movie$|$Star Wars$|$' is the word-based tag sequence with delimiter `$|$',
and the $pos$ list of the sequence is $[0,1,0,1,2,3,0,1,2]$, in which we regard the delimiter as the last word of the tag.
Such local positional encoding strategy is able to disengage tags and tightly
coupling the words from a same tag. The effect of our local positional encoding will be assessed in \S~\ref{sec:ablastu}.
\subsection{Tag Sequence Reordering}
\label{sec:ordering}

The decoder of our model generates each tag by taking all previous outputs as conditions.
Thus the order of the tag sequence would affect the tag generation result.
However, in tag recommendation, the orders of the tag sequences are usually given randomly, which makes it difficult to train the model and cannot fully leverage the semantic relations among tags. To alleviate this, we propose two rules for sorting the tags by accounting for the semantics.

Actually, the tags relevant to a specific source text usually have hierarchical relationships on semantic. For example, the tag ``\textit{Rogue One: A Star Wars Story}'' can be seen as one of the subtopics of the tag ``\textit{Star Wars}''. We further find that the frequency of the tags with abstract meaning is higher than that of the tags representing specific object,
especially in the topics with semantic overlaps (as shown in Table~\ref{tab:tagfreq}).
Therefore, it can be reasonably assumed that the higher the frequency of the labels, the more general the meaning they represent.
Based on this essential assumption,
we propose to reorder the tag sequence in ascending order of frequency (\textit{Order 1}) and in descending order of frequency (\textit{Order 2}) respectively.
\begin{table}[t]
  \centering
  \begin{tabular}{c|| cc }
  \hline\thickhline
  Tag's Name    &Rank &\#Freq. \\
  \hline
  Psychology   &2 &15,171 \\
  Psychic Trauma &6,107 &48\\
  Post-traumatic Stress Disorder &12,613 &17 \\
  \hline
  Movie &3 &15,116 \\
  Star Wars &215 &1,743 \\
  Rogue One: A Star Wars Story &12,287 &18\\
  \hline
  \end{tabular}
  \caption{\textbf{Statistics regarding to the frequency of tags in \textit{Zhihu} dataset,} showing that the tags with more abstract meanings tend to appear in the dataset more frequently.
  }
  \label{tab:tagfreq}
\end{table}
%
%
%

\noindent\textit{\textbf{Order 1.}}
For label generation, given a specific topic, it is easier to predict more abstract topics.
For example, if the model has generated the tag ``\textit{Rogue One: A Star Wars Story}'', then, it is naturally to infer ``\textit{Star Wars}'' and further, the tag of ``\textit{science fiction film}''.
The ambiguity of this process is relatively small compared to the opposite process.

\noindent\textit{\textbf{Order 2.}}
For \textit{Order 2}, we assume that the tags with lower frequency are derived from high-frequency tags.
So the decoder is trained to generate high-frequency tags preferentially and then predict less frequent tags based on previous outputs.
In this way, first the model determines the macro topic of the input, then the previous output will gradually decrease the search scope of the decoder step by step.
Fig.~\ref{fig:introduction1} shows an example of two sets of tags relevant to two questions in \textit{Zhihu}, wherein ``movie'' is the most prior tag followed by ``science fiction movie'' and ``Star Wars'' for the two instances respectively.

The above strategies offer two intuitive yet effective decoding rules from different perspectives, which will be fully evaluated in \S~\ref{sec:tagorder}.

\section{Experiment}
\label{sec:exp}
\subsection{Experimental Setup}
\label{sec:exp:dataset}
\noindent\textbf{Datasets.} In our experiments, following two datasets: \textit{Zhihu} and \textit{Weibo} are used for thoroughly accessing the performance of the proposed approach. The characteristics of these two datasets are summarized in Table~\ref{tab:dataset} and Table~\ref{tab:dataset_label}.

\textit{Zhihu} is from a share task in NLPCC 2018: Automatic Tagging of Zhihu Questions\footnote{\url{http://tcci.ccf.org.cn/conference/2018/taskdata.php}}.
It is a collection of questions in the community question answering web site Zhihu\footnote{\url{https://www.zhihu.com/}}, each of which contains a title, a set of relevant tags and an additional description.
We combine the question title text and the description text into one sentence as the input for all models described below.
Each tag is labeled collaboratively by users from Zhihu.
Since the labels of the test dataset in \textit{Zhihu} are not released, we crawled the corresponding labels from the Zhihu website.

\textit{Weibo} is a corpus of microblogs in Sina Weibo\footnote{\url{https://weibo.com/}} downloaded from NLPIR 5 million microblog corpus\footnote{\url{http://www.nlpir.org/download/weibo.7z}}.
It contains nearly 5 million microblogs, some of which contain tags between two ``\#'' labeled by users of Sina Weibo.
%
%
\begin{table}[t]
  \centering
  \begin{tabular}{m{1.3cm}<{\centering}||p{1.3cm}<{\centering} p{1.3cm}<{\centering} p{1.3cm}<{\centering}}
  \hline\thickhline
  Datasets    &\#Train &\#Dev. &\#Test \\
  \hline
  \textit{Zhihu} &721,608 &8,946 &20,596 \\
  \textit{Weibo} &441,366 &10,000 &10,000 \\
  \hline
  \end{tabular}
  \caption{\textbf{Statistics of \textit{Zhihu} and \textit{Weibo} datasets} regarding to the number of question-tag and microblog-hashtag  pairs.}
  \label{tab:dataset}
\end{table}
\begin{table}[t]
  \centering
  \begin{tabular}{m{1.3cm}<{\centering}||p{1.3cm}<{\centering} p{1.3cm}<{\centering} p{1.3cm}<{\centering}}
  \hline\thickhline
  Datasets &\#Total &\#Average &\#Words\\
  \hline
  \textit{Zhihu} &25,551 &3.13 &2.19\\
  \textit{Weibo} &13,426 &1.05 &3.14\\
  \hline
  \end{tabular}
  \caption{\textbf{Statistics of \textit{Zhihu} and \textit{Weibo} datasets regarding to the number of tag labels}.
  ``\#Total'' is the total individual labels number;
  ``\#Average'' is the average label number that every item in the dataset contains;
  ``\#Words'' is the average word number per labels.}
  \label{tab:dataset_label}
\end{table}

\noindent\textbf{Data Preprocessing.}
We remove the URLs from the input data for all datasets.
The Chinese part for each corpus is segmented by the LTP~\cite{che2010ltp} Chinese word segmentor.
The Chinese examples are presented in segmented romanized form and followed by the English translation in parentheses.

\noindent\textbf{Implementation Details.}
We implement a tag-based model (L2A-label) and a word-based model (L2A-word) for comparison.
The model consists of a 2-layer encoder and a 4-layer decoder,
For the decoder, we use the proposed local positional encoding instead of the ordinary positional encoding~\cite{vaswani2017attention}.
The input vocabulary consists of 80,000 words.
The target side vocabulary contains all tag names and all the words that make up the tags for tag-based (L2A-label) and word-based (L2A-word) decoder respectively.
Specifically, the vocabulary sizes of our word-based decoder are 18,966 and 10,860 including ``tag delimiter'', ``padding'' and ``end of sequence'' symbols for Zhihu and Weibo datasets, respectively.

The beam search and \textit{N}-best voting are adopted to optimize the performance,
which will be detailed below. The beam size and the \textit{N} of \textit{N}-best voting are set to 48 unless otherwise specified.

\noindent\textbf{Baselines.} We compare the proposed model with three baselines:
(\romannumeral1) \textit{CNN}~\cite{gong2016hashtag}: a convolutional neural network based multi-label text classifier;
(\romannumeral2) \textit{LSTM}~\cite{li2016tweet}: a bidirectional LSTM based multi-label text classifier;
and (\romannumeral3) \textit{Topical Attention}~\cite{li2016hashtag}: an attention-based bidirectional LSTM multi-label text classifier with topic model;

For LSTM-based and CNN-based models, the dimension of the hidden states and the number of convolution kernels are set to 512.
And classes in top-5 scores are used as the predicted tags.

\noindent\textbf{Evaluation Metrics.} All experimental results are evaluated on positional weighted precision ($P$), recall ($R$) and $F_1$-score ($F_1$) using the evaluating script task6\_eval.py\footnote{\url{http://tcci.ccf.org.cn/conference/2018/dldoc/tasktestdata06.zip}} provided by NLPCC 2018 share task6.

\subsection{Performance on Multi-Label Classification}
\label{multiclass}
We further evaluate our approach on multi-label classification task using \textit{Zhihu} and \textit{Weibo} datasets. There is no any unseen tag in the test sets of these two datasets.
The average number of tags in \textit{Weibo} is close to 1, which means that for most instance, there is only one relevant tag.
Therefore, we do not perform tag reordering in \textit{Weibo}.
Experimental results on these two datasets are listed in Table~\ref{tab:zhihu} and Table~\ref{tab:weibo} respectively.
We can find that the proposed sequence-to-sequence based model outperforms other traditional sentence classification methods across all the datasets.
We attribute this to the efficient modeling of the semantic relations and the use of word-level cues encoded in the tag sentences.
In next section,
different ingredients and variants of our method will be studied to give a deeper insight into our model design.
%
\begin{table}[t]
  \centering
  \setlength\tabcolsep{5pt}
  \begin{tabular}{c|| ccc | ccc}
  \hline\thickhline
  \multirow{2}{*}{Methods}
  &\multicolumn{3}{c}{Dev} &\multicolumn{3}{c}{Test} \\
  \cline{2-7}
                &$P \uparrow$  &$R \uparrow$  &$F_1 \uparrow$ &$P \uparrow$  &$R \uparrow$  &$F_1 \uparrow$   \\
  \hline
  \hline
  CNN~\cite{gong2016hashtag}   &32.9 &46.8    &38.6 &27.6 &40.2 &32.7\\   

  LSTM~\cite{li2016tweet}     &35.1 &50.0    &41.3  &31.4 &45.8 &37.3   \\ 
  Topical Attention~\cite{li2016hashtag}  &38.1 &\textbf{54.3} &44.7 &34.8 &\textbf{47.2} &40.1 \\
  \hline
  \hline
  L2A-label (\textit{Random}) &40.8 &42.9 &41.8 &36.6 &40.1 &38.2\\ 
  \textit{Order 1}       &44.0 &43.2 &43.5 &40.8&39.5&39.9\\ 
  \textit{Order 2}       &42.6 &42.7 &42.7 &38.9 &38.7 &38.8\\ 
  L2A-word  (\textit{Random})  &43.2 &38.7 &40.8 &37.5 &35.3 &36.4\\
  \textit{Order 1}       &\textbf{47.0} &42.9 &\textbf{44.9} &\textbf{43.3} &38.7 &\textbf{40.9}\\
  \textit{Order 2}       &43.7 &43.5 &43.6 &38.5 &41.0 &39.7\\
  \hline
  \end{tabular}
  \caption{\textbf{Experimental results on \textit{Zhihu} dataset.} 
  }
  \label{tab:zhihu}
\end{table}

\begin{table}[t]
  \centering
  \setlength\tabcolsep{4pt}
  \begin{tabular}{c|| ccc | ccc}
  \hline\thickhline
  \multirow{2}{*}{Methods}
  &\multicolumn{3}{c}{Dev} &\multicolumn{3}{c}{Test} \\
  \cline{2-7}
                &$P  \uparrow$  &$R  \uparrow$ &$F_1  \uparrow$ &$P  \uparrow$  &$R  \uparrow$  &$F_1  \uparrow$   \\
  \hline
  \hline
  CNN~\cite{gong2016hashtag}      &53.9 &51.5    &52.7  &51.1 &48.8 &49.9   \\
  LSTM~\cite{li2016tweet}         &54.7 &54.2 &54.5  &52.2 &52.4 &52.3   \\
  Topical Attention~\cite{li2016hashtag}   &54.3 &\textbf{55.3} &54.7  &52.0 &\textbf{52.9} &52.4   \\
  \hline
  \hline
  L2A-label     &53.1 &52.2 &52.6  &54.3 &49.9 &52.0   \\
  L2A-word      &\textbf{57.0} &54.5 &\textbf{55.7}  &\textbf{56.1} &50.7 &\textbf{53.3}   \\
  \hline
  \end{tabular}
  \caption{\textbf{Experimental results on \textit{Weibo} dataset}; showing that the suggested model outperforms previous text classification based methods. See \S~\ref{multiclass} for more details.
  }
  \label{tab:weibo}
\end{table}

\subsection{Generating Unseen Tags}
\label{sec:unseentags}
%

Previous classification based tag recommendation methods pre-defined a fixed number of tags.
Differently, the core idea of the proposed approach lies on treating the tag recommendation as word-level text generation task.
It is freed from the limitation that the predicted tags should have to already exist in the training dataset.
The suggested model is capable to generate unseen new tags, which is more favored than previous multi-label classification based methods.

To evaluate the ability of the model to generate new tags, we introduce an additional test set, Test-open, which is automatically crawled from \textit{Zhihu} web site and mainly includes the topics of ``college'', ``life'' and ``recommendation''.
Test-open contains 4,000 \textit{Zhihu} questions and 13,165 tags in total.
Every instance in the dataset contains at least one new tag and
the total ratio of new tags is 38.52\%.
%
%
The experimental results are shown in Table~\ref{tab:unseen}.
It can be observed that our word-based method (\textit{L2A}) can generate new tags successfully, as we treat the tag recommendation as a word-level text generation task.
\begin{table}[t]
  \centering
  \setlength\tabcolsep{4pt}
  \begin{tabular}{c||cccc}
  \hline\thickhline
  Models    &$P \uparrow$ &$R \uparrow$ &$F_1 \uparrow$ &\#new $\uparrow$\\
  \hline
  \hline
  CNN~\cite{gong2016hashtag} &19.6 &19.1 &19.4 &-\\
  LSTM~\cite{li2016tweet} &21.2 &18.7 &19.9 &-\\
  Topical Attention~\cite{li2016hashtag} &22.5 &19.0 &20.6 &-\\
  \hline
  \hline
  L2A-word (\textit{Random}) &21.2 &20.1 &20.6 &53\\
   \textit{Order 1}&24.6 &17.8 &20.7 &23\\
   \textit{Order 2}&\textbf{25.0} &\textbf{23.4} &\textbf{24.2} &\textbf{82}\\
  \hline
  \end{tabular}
  \caption{\textbf{Experimental results on Test-open,} where
  ``new'' is the number of correct new tags which are generated by the corresponding model. It shows that the proposed model gains promising results, especially for unseen tags. See \S~\ref{sec:unseentags} for more details.
  }
  \label{tab:unseen}
\end{table}

\begin{table}[ht]
  \centering
  \renewcommand\arraystretch{1.}
  \setlength\tabcolsep{9pt}
  \newcommand{\tabincell}[2]{\begin{tabular}{@{}#1@{}}#2\end{tabular}}
  \begin{tabu}{c||c}
  \hline\thickhline
  Src
  &\tabincell{c}{
  \textit{zai {\color{blue}\underline{hebei shifandaxue}} shangxue shi yi zhong zenyang de tiyan?} \\
   (How do you feel about studying at \underline{Hebei Normal University}?)} \\
  \hline
  Ref
  &\tabincell{c}{ 
  1.\textit{daxue}~(college),~~\\
  2.\textit{hebeisheng}~(Hebei province),\\
  3.\textit{tiyan lei wenti}~(questions about experience),\\
  4.\textit{\color{blue}\underline{hebei shifandaxue}}~(Hebei Normal University)
  } \\
  \hline
  \tabincell{c}{Topical Attention\\\cite{li2016hashtag}}
  &\tabincell{c}{ 
  1.\textbf{\textit{hebeisheng}~(Hebei province)},~~
  2.\textbf{\textit{daxue}~(college)},~~\\
  3.\textit{renji jiaowangn}~(interpersonal communication),\\
  4.\textit{daxue shenghuo}~(college life),~~\\
  5.\textit{shifandaxue}~(normal university)
  }\\
  \hline
  \tabincell{c}{L2A-word (\textit{Random})}
  &\tabincell{c}{
  1.\textit{daxue jiudu tiyan}~(experience of studying in college),\\
  2.\textit{daxue shenghuo}~(college life),~~\\
  3.\textbf{\textit{daxue}~(college)},~~\\
  4.\textit{shifandaxue}~(Normal University),\\
  } \\
  \hline %
  \tabincell{c}{L2A-word (\textit{Order 1})}
  &\tabincell{c}{
  1.\textit{shifandaxue}~(Normal University),\\
  2.\textit{daxue jiudu tiyan}~(experience of studying in college),\\
  3.\textbf{\textit{daxue}~(college)},~~\\
  3.\textbf{\textit{hebeisheng}~(Hebei province)}\\
  } \\
  \hline %
  \tabincell{c}{L2A-word (\textit{Order 2})}
  &\tabincell{c}{ 
  1.\textbf{\textit{daxue}~(college)},~~\\
  2.\textbf{\textit{hebeisheng}~(Hebei province)},~~\\  
  3.\textbf{\textit{{\color{blue}\underline{hebei shifandaxue}}}~(Hebei Normal University)}
  } \\
  \hline %
  \end{tabu}
  \caption{\textbf{Examples of generating unseen tags.} ``Src'' indicates the input. ``Ref'' is the manually labeled tag.
  The \textbf{bold phrases} are correct predicted tags;
  The {\color{blue}\underline{underlined phrases}} in ``Src'' is the potential tag's name that  never appeared in the training corpus. Above examples show that our sequence-to-sequence model (L2A) is able to produce more accurate tag recommendations compared with previous text classification based models, and handles unseen tags (``\textit{hebei shifandaxue}'') well. See \S~\ref{sec:unseentags} for more details.
  }
  \label{tab:unseenexp}
\end{table}

Table~\ref{tab:unseenexp} shows a detailed example of unseen tag generation, where our word-based model (with \textit{Order 2}) predicts a new tag ``\textit{hebei shifandaxue}'' (Hebei Normal University) successfully. 
By applying word embeddings, our model can effectively capture the semantic relation between tag-words (``\textit{hebei}'' and ``\textit{beijing}'', a city and a province of China, respetively).
This demonstrates that our model is able to learn general patterns from training corpus and further leverages such cues to generate unseen tags.
\subsection{Tag Order}
\label{sec:tagorder}
We assess the effect of using different ordering methods in \S~\ref{sec:ordering}.
The results are summarized in Table~\ref{tab:unseen} and Table~\ref{tab:zhihu}.
For unseen tags generation task (Table~\ref{tab:unseen}), the model with \textit{Order 2} outperforms the one with \textit{Order 1}, which is contrary to the results for multi-class classification (Table~\ref{tab:zhihu}).
As mentioned in \S~\ref{sec:ordering}, \textit{Order 2} first predicts the tags with high frequency and then infers the rare tags.
Since high-frequency tags are usually abstract and more common in the corpus, they are more likely to be correctly predicted,
but the following tag generation step may be ambiguous.
\textit{Order~1} first predicts rare tags which often have concrete meanings.
The prediction from concrete tags to abstract ones is usually unambiguous.
Therefore, if the test set is out of domain and contain lots of unseen tags, first generating high-frequency may be safer (case of \textit{Order 2} in Table~\ref{tab:unseen}).
On the contrary, if the test set share similar distribution with the training corpus (case in Table~\ref{tab:zhihu}), \textit{Order 1} is more favored.

From Table~\ref{tab:unseen} and Table~\ref{tab:zhihu}, we can see that the models with \textit{Order~1} gain higher precision score rather than recall rate,
suggesting the certainty of \textit{Order~1}.
On the other hand, the models with \textit{Order~2} have higher recall, verifying that \textit{Order~2} could assign the model with a wider searching space.
In addition, the models equipped with reordering strategies, either \textit{Order~1} or \textit{Order~2}, outperform their counterparts with random order,
proving the effectiveness of the proposed ordering strategies.

\subsection{\textit{N}-Best Voting.}
\label{sec:nbestvoting}
The greedy inference strategy may lead the generated tags to focus on a specific topic.
If the first predicted label is wrong, it is very likely that the whole sequence cannot be predicted correctly~\cite{wang2016cnn}.
We apply beam search algorithm to the decoder and propose a voting-based label screening method called \textit{N}-best voting
to make full use of the \textit{N}-best list and reduce the impact of accidental translation errors.
More specially, in this method, we count the total frequency of tags in the \textit{N}-best list and
select the tags with the frequency higher than a threshold.
The voting method gains the best performance on $F_1$ score in the classification task when the threshold is set to one-fourth of the beam size.
\begin{figure}[ht]
\centering
\includegraphics[width=0.6\textwidth]{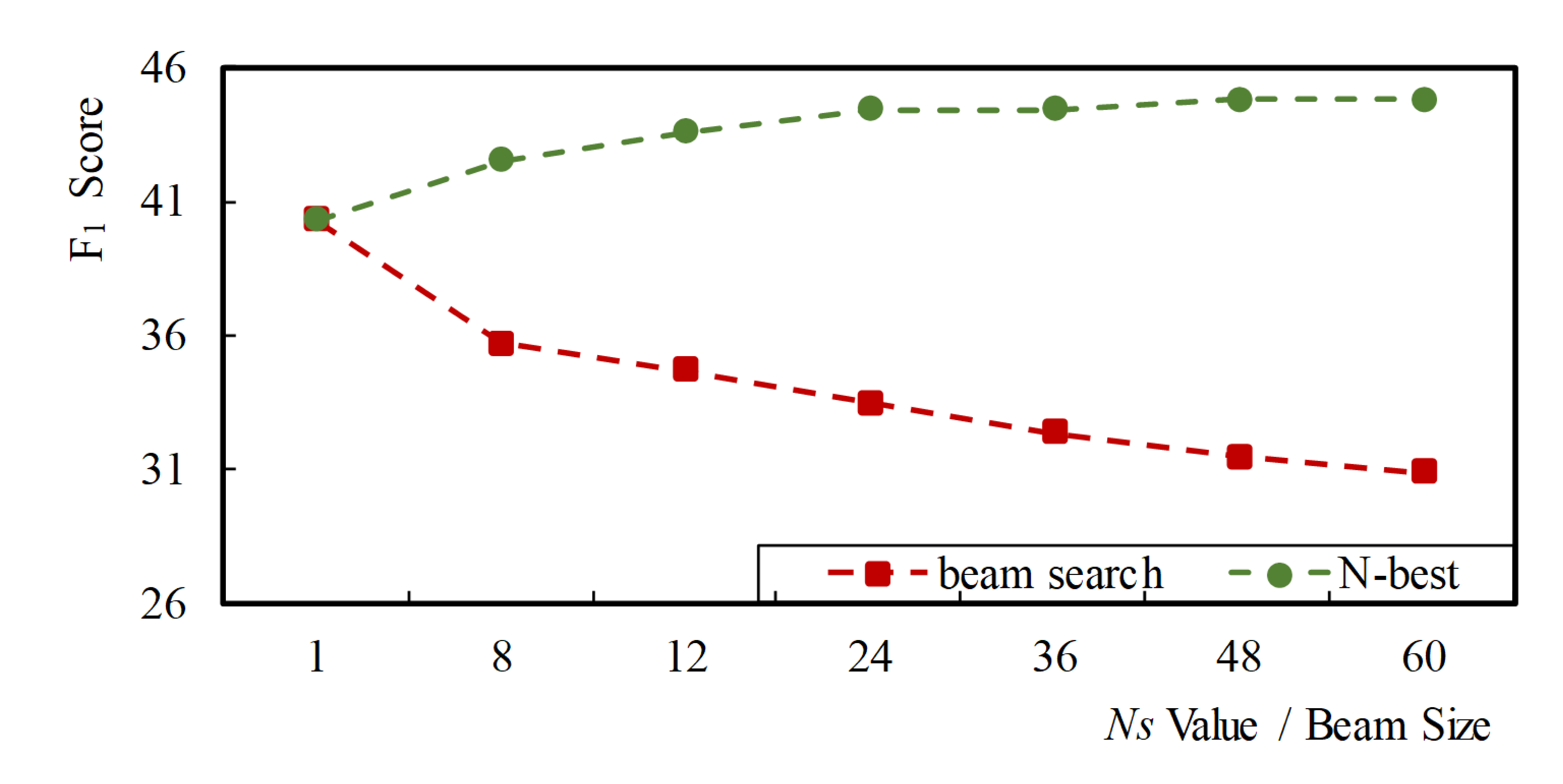}
\caption{\textbf{Performance comparison between using beam search only and adopting $\textit{N}$-best voting in different beam sizes and $N$s}; showing the performance improvement after using $\textit{N}$-best voting. See \S~\ref{sec:nbestvoting} for details.
}
\label{fig:nbest_voting}
\end{figure}
Fig.~\ref{fig:nbest_voting} shows that the model with \textit{N}-best voting performs obviously better than the one using beam search only.
It also shows the performance raises with the increase of the size of \textit{N}-best list.

\subsection{Unseen tags vs unseen meaningless tags.}
\label{sec:meaningless}
The word-based model is trained to generate word sequence with tag's name delimiters in it without explicitly ``knowing'' the full tag's name.
In other words, there are no constraint for the integrity of tags when combing tag-words.
This allows our model the capacity of generating unseen tags, while also introduces the potential risk of generating unseen meaningless tags (error tags).
Therefore, it is interesting to explore influence of those error tags.
Our N-best voting reduces the potential risk of producing unseen meaningless tags,
because it considers the more confident tags (with higher frequency).
Our reordering strategies also partially remedy this, as they regularize the flexible generation space of our seq2seq model.

Table~\ref{tab:translationerrors} shows the number of error tags and the error rates generated by the model with \textit{Order 1} on the Dev. data of \textit{Zhihu} datasets.
From Table~\ref{tab:translationerrors}, we can see that meaningless tags rarely appear in the generated tag sequences, and using the \textit{N}-best voting strategy can reduce the error rates effectively.
In Table~\ref{tab:translationerrors}, 1,076,588 in \#Outputs column is the total number of \textit{N}-best (\textit{N} = 48) outputs of our model,
so it is significantly larger than the number of the input Zhihu questions (8,946) and with many wrong predictions (11,814).
After using \textit{N}-best voting, most unconfident predictions are removed, and thus achieving a very low error rate (0.03\%).

\begin{table}[ht]
  \centering
  \setlength\tabcolsep{8pt}
  \begin{tabular}{c||ccc}
  \hline\thickhline
  Models &\#Meaningless &\#Outputs  &Error Rate(\%)  \\
  \hline
  \hline
  \textit{Order 1} &11,814 &1,076,588  &1.10 \\
  +\textit{N}-Best voting &9 &26,444  &0.03\\ 
  \hline
  \end{tabular}
  \caption{\textbf{Translation errors generated by L2A-word};
  showing that our method is robust to generated errors and the \textit{N}-best voting strategy is able to decrease the error rate efficiently.
  \#Meaningless is the number of generated errors;
  \#Outputs is the number of total number of \textit{N}-best (\textit{N} = 48) outputs;
  and \#Error Rate is the percentage of \#Meaningless in \#outputs.
  See \S~\ref{sec:meaningless} for details.}
  \label{tab:translationerrors}
\end{table}

%
\subsection{Attention-based Decoder}
\label{sec:attention}
\begin{figure}[t]
\centering
\subfigure[Unseen Tag]
{
  \begin{minipage}{1\textwidth}
  \centering
  \includegraphics[width=0.7\textwidth]{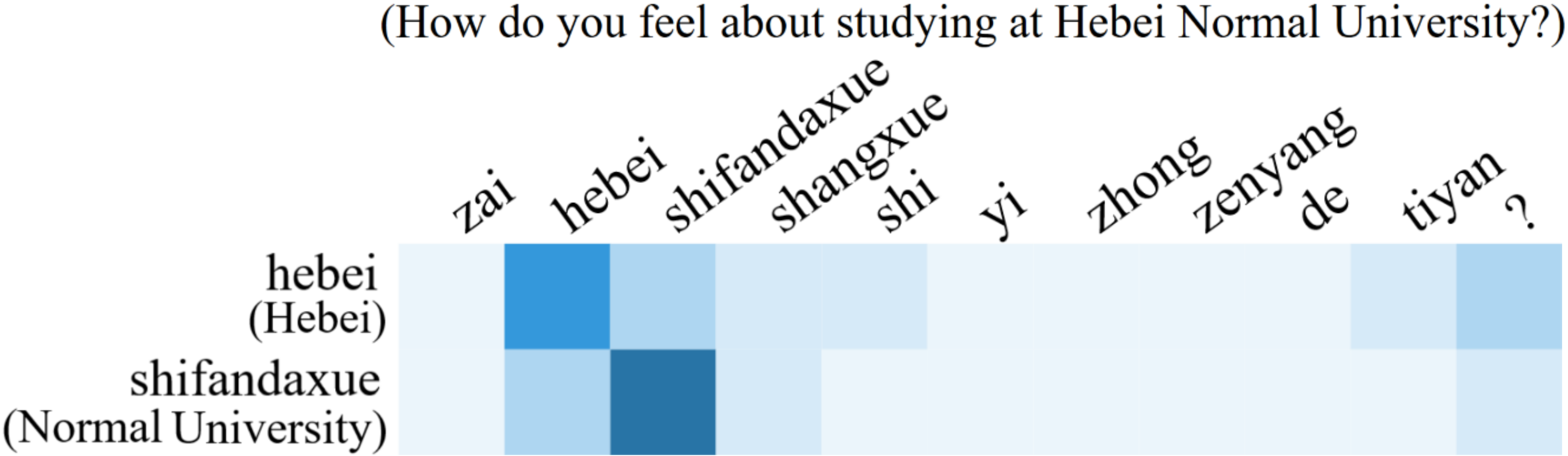}
  \end{minipage}
}
\subfigure[Multi-label Classification]
{
  \begin{minipage}{1\textwidth}
  \centering
  \includegraphics[width=0.71\textwidth]{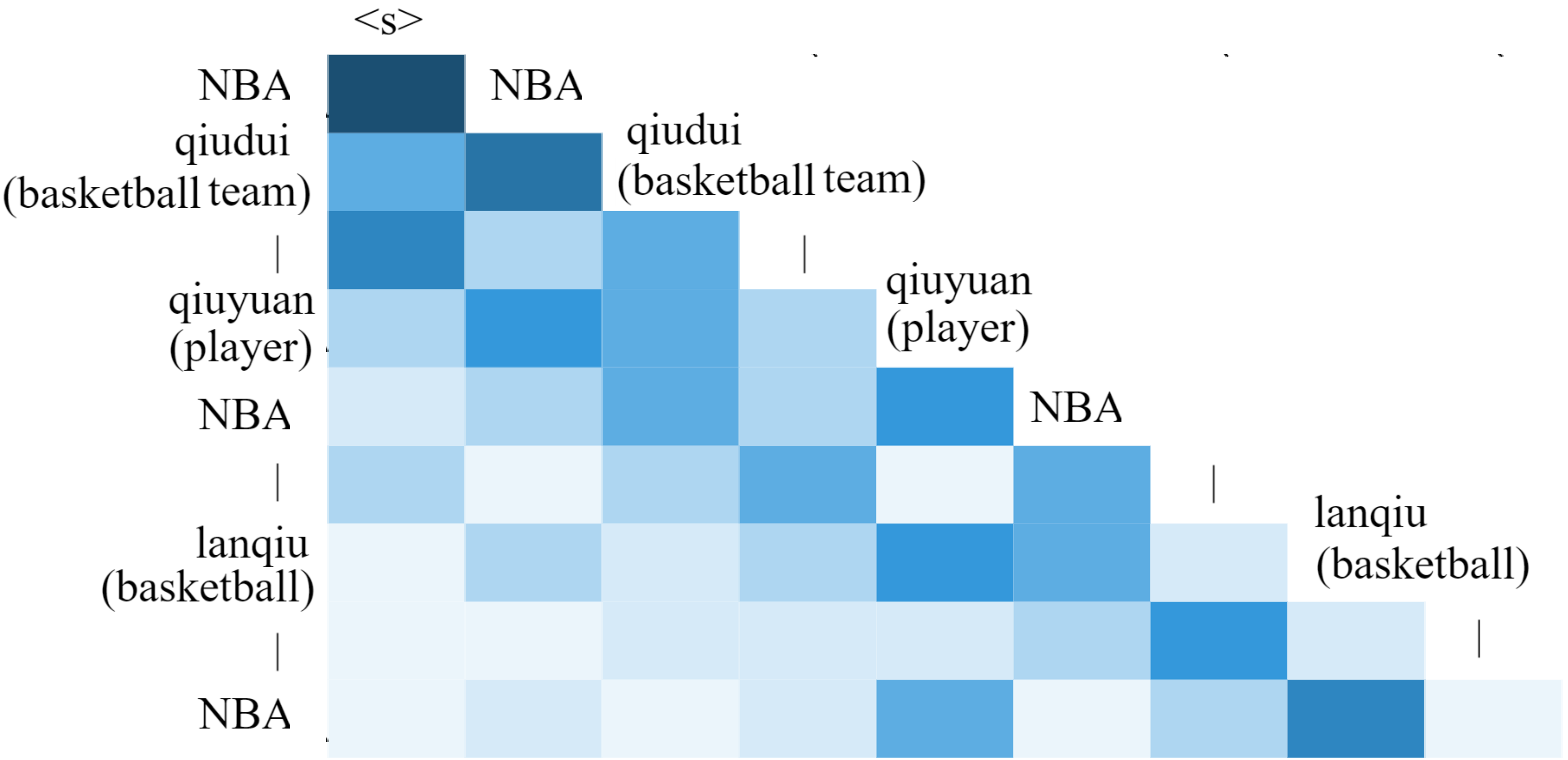}
  \end{minipage}
}
\caption{\textbf{Can the attention based decoder capture the sematic relations?}
(a) the source inputs  or (b) pre-generated tags  lie on the x-axis, and the new predicted tag lies on the y-axis.
Corresponding English Translations are noted in bracket.
Darker colors represent higher attention weights.
In (a), ``\textit{hebei shifandaxue}'' is an unseen tag, but it has been successfully predicted by considering the related tag-words ``\textit{hebei}'', ``\textit{shifandaxue}'' in the source input.
(b) shows that, when predicting ``\textit{qiuyuan}'', ``\textit{qiudui}'', ``\textit{lanqiu}'' and ``\textit{NBA}'' labels, our model orients attention to those semantically related tags which are generated previously.
The word ``$\langle s \rangle$'' is the start symbol of the tag sequence which is the given input of the decoder at the initial time step.
Above two example shows that the proposed model is able to leverage the word-level cues for inferring unseen tags (a) and capture the rich semantic relations between tags (b). See \S~\ref{sec:attention} for more details.
}
\label{fig:attention}
\end{figure}

To give an in-depth analysis of the proposed attention-based decoder,
we show in Fig.~\ref{fig:attention} two visual examples of the attentions learned by the Transformer decoder.
The source inputs/pre-generated tags and the newly predicted tags are shown on the x-axis and the y-axis, respectively.
Fig.~\ref{fig:attention}~(a) shows the ``\textit{hebei shifandaxue}'' is generated by the model with \textit{Order 2} mainly considering the cues ``\textit{hebei}'', ``\textit{shifandaxue}'' in the source input.
In Fig.~\ref{fig:attention}~(b), it gives an example of the decoder self-attention weights of the model with \textit{Order 1}.
The input question of the example is \textit{``Zhouqi qianyue qu Xinjiang he Durant hen leisi a, weisha dangchu zhiyi de ren name shao?''} 
(Why are there few people who have disputes about Zhouqi¡¯s signing of the Xinjiang?).
From Fig.~\ref{fig:attention}~(b) we can see that the newly predicted tag-words, such as ``\textit{lanqiu}'' (basketball) and ``NBA'' are mostly influenced by the semantically related words (cues) in the prior predicted tags, such as ``NBA'', ``\textit{qiuyuan}'' (player) and ``\textit{lanqiu}'' (basketball).
Above two example shows that the proposed model is able to leverage the word-level cues for inferring unseen tags
(Fig.~\ref{fig:attention}~(a)) and capture the rich semantic relations between tags (Fig.~\ref{fig:attention}~(b)).

\subsection{Ablation Study}
\label{sec:ablastu}
In this section, we analyze the contribution of the model components in the final performance and give a more in-depth insight into our model design via comparison to several variants.

\noindent\textbf{Local Positional Encoding.}
Next we study the effectiveness of the proposed local position encoding strategy in \S~\ref{sec:ordering}.
We compare the performance of our model \textit{w.} and\textit{ w/o.} positional encoding~\cite{vaswani2017attention} and our model \textit{w.} local positional encoding on \textit{Zhihu} Dev. dataset.
The comparison results are reported in Table~\ref{tab:local_position_encoding}.
We find that the model (\textit{w/.} local p.e.) gains the best performance, demonstrating the effectiveness of the proposed local positional encoding strategy.

\begin{table}[ht]
  \centering
  \setlength\tabcolsep{8pt}
  \begin{tabular}{c||ccc}
  \hline\thickhline
  Models    &$P  \uparrow$ &$R  \uparrow$ &$F_1  \uparrow$ \\
  \hline
  \hline
  \textit{w/.} local p.e. (Ours) &\textbf{47.0} &\textbf{42.9} &\textbf{44.9}\\
  \hline
  \textit{w/o.} p.e. &$46.4_{{\color{blue}{(\textbf{-0.6})}}}$ &$42.1_{{\color{blue}{(\textbf{-0.8})}}}$ & $44.1_{{\color{blue}{(\textbf{-0.8})}}}$\\
  \textit{w/.}  p.e. &$46.0_{{\color{blue}{(\textbf{-1.0})}}}$ &$42.6_{{\color{blue}{(\textbf{-0.3})}}}$ &$44.2_{{\color{blue}{(\textbf{-0.7})}}}$\\
  \hline
  \end{tabular}
  \caption{\textbf{Comparison of using different positional encoding methods on \textit{Zhihu} Dev. dataset.}
  ``p.e.'' is the abbreviation of ``positional encoding''. The
relative performance change is reported in {\scriptsize${{\color{blue}{(\cdot)}}}$}.  See \S~\ref{sec:ablastu} for details.
  }
  \label{tab:local_position_encoding}
\end{table}
\noindent\textbf{Model Architecture.} To fully assess the proposed LSTM-Attention model, we consider serval variants. For the sequence-to-sequence architecture, we adopt LSTM and Transformer based attention model as their implementations. Thus we have four variants derived from the different combinations of encoder/decoder structures: L2L, A2A, A2L, and L2A. Here
``L''  and ``A'' are the abbreviations of ``LSTM'' and ``Attention Model (Transformer~\cite{vaswani2017attention})'' respectively.
For example, ``L2A'' represents the proposed LSTM-Attention model that uses LSTM as the encoder and Attention Model as the decoder.
The performance of above variants on  \textit{Zhihu} Dev. dataset is summarized in Table~\ref{tab:model_architecture}.
It can be observed that the proposed LSTM-Attention model (L2A) outperforms other variants, since the LSTM based encoder can well capture the context information of the input sequence and the Transformer based decoder is suitable for the situation that the order of tag sequence is not strict.

\begin{table}[ht]
  \centering
  \setlength\tabcolsep{10pt}
  \begin{tabular}{c||ccc}
  \hline\thickhline
  Models    &$P  \uparrow$ &$R  \uparrow$ &$F_1  \uparrow$ \\
  \hline
  \hline
  %
  L2A (Ours) &\textbf{47.0} &\textbf{42.9} &\textbf{44.9}\\
  \hline
  L2L &$36.7_{{\color{blue}{(\textbf{-10.3})}}}$ &$42.6_{{\color{blue}{(\textbf{-0.3})}}}$ &$39.4_{{\color{blue}{(\textbf{-5.5})}}}$ \\
  A2A &$40.1_{{\color{blue}{(\textbf{-6.9})}}}~~$  &$41.5_{{\color{blue}{(\textbf{-1.4})}}}$ &$40.8_{{\color{blue}{(\textbf{-4.1})}}}$ \\
  A2L &$37.1_{{\color{blue}{(\textbf{-9.9})}}}$ &$40.1_{{\color{blue}{(\textbf{-2.8})}}}$ &$38.5_{{\color{blue}{(\textbf{-6.4})}}}$ \\
  \hline
  \end{tabular}
  \caption{\textbf{Comparison of different combinations of encoder and decoder on \textit{Zhihu} Dev. dataset}; showing the proposed LSTM-Attention model (L2A) is more favored.  The
relative performance change is reported in {\scriptsize${{\color{blue}{(\cdot)}}}$}.  See \S~\ref{sec:ablastu} for details.
  }
  \label{tab:model_architecture}
\end{table}

\section{Conclusions}
We propose a novel word-based tag recommendation method, which tackles tag recommendation as word-based tag-sequence generation.
Our approach is achieved by a carefully designed LSTM-Attention model, which is able to effectively capture the rich semantic relations and sequential dependencies within tag sequences.
To better disentangle tags and compact the words from the same tags, we extend the Transformer based decoder with a local positional encoding strategy. In addition, two tag ordering methods are proposed for better leveraging the semantic relations.
Experimental results show our model outperforms other state-of-the-art multi-class classifier based tag recommendation models and is flexible to generate unseen tags.
In the future, we will introduce pointer networks~\cite{vinyals2015pointer} to find the tags within the inputs more directly
and explore to integrate knowledge base into our approach for modeling the semantics of tags more explicitly.

\bibliography{dasfaa2019}

\begin{thebibliography}{10}
\providecommand{\url}[1]{\texttt{#1}}
\providecommand{\urlprefix}{URL }
\providecommand{\doi}[1]{https://doi.org/#1}

\bibitem{bansal2015towards}
Bansal, P., Jain, S., Varma, V.: Towards semantic retrieval of hashtags in
  microblogs. In: Proceedings of the 24th International Conference on World
  Wide Web Companion, {WWW} 2015, Florence, Italy, May 18-22, 2015 - Companion
  Volume. pp.~7--8 (2015)

\bibitem{buddhitha2015topic}
Buddhitha, P., Inkpen, D.: Topic-based sentiment analysis. In: Information
  Management and Big Data - Second Annual International Symposium, SIMBig 2015,
  Cusco, Peru, September 2-4, 2015, and Third Annual International Symposium,
  SIMBig 2016, Cusco, Peru, September 1-3, 2016, Revised Selected Papers. pp.
  95--107 (2016)

\bibitem{che2010ltp}
Che, W., Li, Z., Liu, T.: {LTP:} {A} chinese language technology platform. In:
  {COLING} 2010, 23rd International Conference on Computational Linguistics,
  Demonstrations Volume, 23-27 August 2010, Beijing, China. pp. 13--16 (2010)

\bibitem{cho2014learning}
Cho, K., van Merrienboer, B., G{\"{u}}l{\c{c}}ehre, {\c{C}}., Bahdanau, D.,
  Bougares, F., Schwenk, H., Bengio, Y.: Learning phrase representations using
  {RNN} encoder-decoder for statistical machine translation. In: Proceedings of
  the 2014 Conference on Empirical Methods in Natural Language Processing,
  {EMNLP} 2014, October 25-29, 2014, Doha, Qatar, {A} meeting of SIGDAT, a
  Special Interest Group of the {ACL}. pp. 1724--1734 (2014)

\bibitem{Efron2010hashtag}
Efron, M.: Hashtag retrieval in a microblogging environment. In: Proceeding of
  the 33rd International {ACM} {SIGIR} Conference on Research and Development
  in Information Retrieval, {SIGIR} 2010, Geneva, Switzerland, July 19-23,
  2010. pp. 787--788 (2010)

\bibitem{gong2013deep}
Gong, Y., Jia, Y., Leung, T., Toshev, A., Ioffe, S.: Deep convolutional ranking
  for multilabel image annotation. arXiv preprint arXiv:1312.4894  (2013)

\bibitem{gong2016hashtag}
Gong, Y., Zhang, Q.: Hashtag recommendation using attention-based convolutional
  neural network. In: Proceedings of the Twenty-Fifth International Joint
  Conference on Artificial Intelligence, {IJCAI} 2016, New York, NY, USA, 9-15
  July 2016. pp. 2782--2788 (2016)

\bibitem{huang2016hashtag}
Huang, H., Zhang, Q., Gong, Y., Huang, X.: Hashtag recommendation using
  end-to-end memory networks with hierarchical attention. In: {COLING} 2016,
  26th International Conference on Computational Linguistics, Proceedings of
  the Conference: Technical Papers, December 11-16, 2016, Osaka, Japan. pp.
  943--952 (2016)

\bibitem{huang2017multi}
Huang, J., Li, G., Wang, S., Xue, Z., Huang, Q.: Multi-label classification by
  exploiting local positive and negative pairwise label correlation.
  Neurocomputing  \textbf{257},  164--174 (2017)

\bibitem{kalamatianos2015sentiment}
Kalamatianos, G., Mallis, D., Symeonidis, S., Arampatzis, A.: Sentiment
  analysis of greek tweets and hashtags using a sentiment lexicon. In:
  Proceedings of the 19th Panhellenic Conference on Informatics, {PCI} 2015,
  Athens, Greece, October 1-3, 2015. pp. 63--68 (2015)

\bibitem{kim2018twitter}
Kim, Y., Hwang, E., Rho, S.: Twitter news-in-education platform for social,
  collaborative, and flipped learning. The Journal of Supercomputing
  \textbf{74}(8),  3564--3582 (2018)

\bibitem{li2016tweet}
Li, J., Xu, H., He, X., Deng, J., Sun, X.: Tweet modeling with {LSTM} recurrent
  neural networks for hashtag recommendation. In: 2016 International Joint
  Conference on Neural Networks, {IJCNN} 2016, Vancouver, BC, Canada, July
  24-29, 2016. pp. 1570--1577 (2016)

\bibitem{li2016hashtag}
Li, Y., Liu, T., Jiang, J., Zhang, L.: Hashtag recommendation with topical
  attention-based {LSTM}. In: {COLING} 2016, 26th International Conference on
  Computational Linguistics, Proceedings of the Conference: Technical Papers,
  December 11-16, 2016, Osaka, Japan. pp. 3019--3029 (2016)

\bibitem{ma2016public}
Ma, B., Yuan, H., Wan, Y., Qian, Y., Zhang, N., Ye, Q.: Public opinion analysis
  based on probabilistic topic modeling and deep learning. Public Opinion
  (2016)

\bibitem{meduru2017opinion}
Meduru, M., Mahimkar, A., Subramanian, K., Padiya, P.Y., Gunjgur, P.N.: Opinion
  mining using twitter feeds for political analysis. International Journal of
  Computer  \textbf{25}(1),  116--123 (2017)

\bibitem{rawat2016contagnet}
Rawat, Y.S., Kankanhalli, M.S.: Contagnet: Exploiting user context for image
  tag recommendation. In: Proceedings of the 2016 {ACM} Conference on
  Multimedia Conference, {MM} 2016, Amsterdam, The Netherlands, October 15-19,
  2016. pp. 1102--1106 (2016)

\bibitem{saha2013discriminative}
Saha, A.K., Saha, R.K., Schneider, K.A.: A discriminative model approach for
  suggesting tags automatically for stack overflow questions. In: Proceedings
  of the 10th Working Conference on Mining Software Repositories. pp. 73--76
  (2013)

\bibitem{vaswani2017attention}
Vaswani, A., Shazeer, N., Parmar, N., Uszkoreit, J., Jones, L., Gomez, A.N.,
  Kaiser, L., Polosukhin, I.: Attention is all you need. In: Advances in Neural
  Information Processing Systems 30: Annual Conference on Neural Information
  Processing Systems 2017, 4-9 December 2017, Long Beach, CA, {USA}. pp.
  6000--6010 (2017)

\bibitem{vinyals2015pointer}
Vinyals, O., Fortunato, M., Jaitly, N.: Pointer networks. In: Proceedings of
  the 28th International Conference on Neural Information Processing
  Systems-Volume 2. pp. 2692--2700. MIT Press (2015)

\bibitem{wang2016cnn}
Wang, J., Yang, Y., Mao, J., Huang, Z., Huang, C., Xu, W.: {CNN-RNN:} {A}
  unified framework for multi-label image classification. In: 2016 {IEEE}
  Conference on Computer Vision and Pattern Recognition, {CVPR} 2016, Las
  Vegas, NV, USA, June 27-30, 2016. pp. 2285--2294 (2016)

\bibitem{wang2016momentum}
Wang, L., Li, X., Liao, L.J., Liu, L.: A momentum theory for hot topic
  life-cycle: A case study of hot hashtag emerging in twitter. International
  Journal of Computers Communications \& Control  \textbf{11}(5),  734--746
  (2016)

\bibitem{yeh2017learning}
Yeh, C., Wu, W., Ko, W., Wang, Y.F.: Learning deep latent space for multi-label
  classification. In: Proceedings of the Thirty-First {AAAI} Conference on
  Artificial Intelligence, February 4-9, 2017, San Francisco, California,
  {USA.} pp. 2838--2844 (2017)

\bibitem{zhang2017hashtag}
Zhang, Q., Wang, J., Huang, H., Huang, X., Gong, Y.: Hashtag recommendation for
  multimodal microblog using co-attention network. In: Proceedings of the 26th
  International Joint Conference on Artificial Intelligence. pp. 3420--3426
  (2017)

\bibitem{zhu2018multi}
Zhu, Y., Kwok, J.T., Zhou, Z.H.: Multi-label learning with global and local
  label correlation. IEEE Transactions on Knowledge and Data Engineering
  \textbf{30}(6),  1081--1094 (2018)

\end{thebibliography}
\bibliographystyle{splncs04}

\end{document}